\documentclass{ecai}            

\usepackage{graphicx}
\usepackage{latexsym}
\usepackage{subcaption}
\usepackage{cite, url}
\usepackage{amsmath}
\usepackage{amssymb}
\usepackage{amsfonts}
\usepackage{algorithmic}
\usepackage{textcomp}
\usepackage{xcolor}
\usepackage{multirow}
\usepackage{threeparttable}
\usepackage{caption} 
\usepackage{float}

\begin{document}

\begin{frontmatter}
\title{FASTC: A Fast Attentional Framework  for Semantic Traversability Classification Using Point Cloud }

\author[a,b]{\fnms{Yirui}~\snm{Chen}}
\author[a,b]{\fnms{Pengjin}~\snm{Wei}\thanks{Equal contribution to the first author.}}
\author[a,b]{\fnms{Zhenhuan}~\snm{Liu}} 
\author[a,b]{\fnms{Bingchao}~\snm{Wang}}
\author[a,b]{\fnms{Jie}~\snm{Yang}}
\author[a,b]{\fnms{Wei}~\snm{Liu}\thanks{This work is supported by Pujiang Program under grant No. 22PJ1406600. Wei Liu and Jie Yang are the corresponding authors of this paper. Email: \texttt{\{weiliucv,jieyang\}@sjtu.edu.cn}}}

\address[a]{\small{Institute of Image Processing and Pattern Recognition, Department of Automation, Shanghai Jiao Tong University, China}}
\address[b]{\small{Key Laboratory of System Control and Information Processing, Ministry of Education, China}}

\begin{abstract} Producing traversability maps and understanding the surroundings are crucial prerequisites for autonomous navigation. In this paper, we address the problem of traversability assessment using point clouds. We propose a novel pillar feature extraction module that utilizes PointNet to capture features from point clouds organized in vertical volume and a 2D encoder-decoder structure to conduct traversability classification instead of the widely used 3D convolutions. This results in less computational cost while even better performance is achieved at the same time.  We then propose a new spatio-temporal attention module to fuse multi-frame information, which can properly handle the varying density problem of LIDAR point clouds, and this makes our module able to assess distant areas more accurately. Comprehensive experimental results on augmented Semantic KITTI and RELLIS-3D datasets show that our method is able to achieve superior performance over existing approaches both quantitatively and quantitatively. Our code is publicly available at \url{https://github.com/chenyirui/FASTC}.
\end{abstract}
\end{frontmatter}

\section{Introduction}
Autonomous driving is a highly challenging task that requires a reliable understanding of the local environment\cite{leonard2008perception}. Producing traversability map is an important part of environment understanding, and the traversability maps can be generated through the point clouds captured with 3D LIDAR sensors.

Although machine learning-based perception has been employed in previous research to enable robots and vehicles to better comprehend their environment, such approaches are primarily tailored for structured urban environments and road networks, limiting their practicality in unstructured off-road rough terrain scenarios. In such environments, it is crucial to comprehend the traversability of the surrounding terrain for effective path planning and decision-making. The goal of traversability estimation is to perceive whether the surroundings are traversable from sparse LIDAR data. Occupancy grids are 2D spatial maps of the environment around the vehicle constructed by LIDAR point cloud data which are often used for path planning {rummelhard2015conditional}. More specifically, traversability estimation aims at finding out the traversability for each of the grid cells, which is a challenging task due to the sparse nature of point clouds. In addition, off-road environments have unstructured class boundaries, uneven terrain, strong textures, and irregular features\cite{JiangOsteen-85}. Addressing these issues would benefit autonomous navigation in complex environments. Despite recent efforts on traversability analysis, the current state-of-the-art approach\cite{shaban2022semantic,choy20194d} faces limitations in terms of speed and accuracy. This is mainly attributed to the following three reasons: Firstly, the utilization of 3D convolution in recent works\cite{choy20194d,shaban2022semantic} causes a high computational cost and memory consumption. Secondly, recurrent neural networks used to aggregate multi-frame information suffer from limitations in capturing significant parts, potentially leading to confusion in the estimation of traversability semantics. Thirdly, semantic networks have limitations in capturing features and may fail to generate accurate predictions due to a restricted receptive field. These limitations hinder the effectiveness of traversability analysis and therefore require further investigation to improve the accuracy and efficiency of the approach.
 
 %------------------------------ Figure ---------------------------------------
 \begin{figure}[!t]
  \centering
  \includegraphics[width=\linewidth]{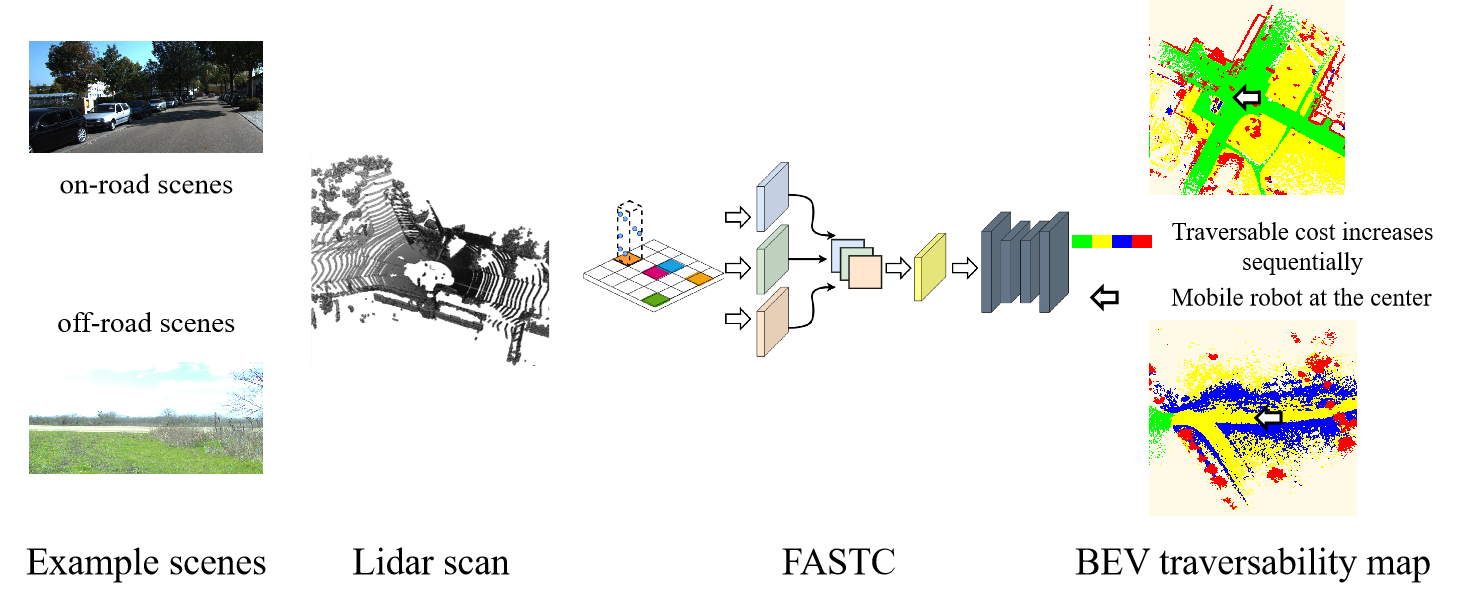}
    \caption{ \textbf{Overview of FASTC. } FASTC is a framework that efficiently generates a semantic traversability map of the surrounding environment by taking point cloud data from 3D LIDAR as input.}
 \label{intro}
\end{figure}
%------------------------------------------------------------------------------

We aim to address the challenge of inferring semantic traversability classes of the environment and locating obstacles rapidly for navigation. Our problem is formulated by following the definition of BEVNet\cite{shaban2022semantic}, which divides the terrain into four cost classes based on the traversability of a vehicle. We take into account various attributes that influence the cost of traversability, including the semantic and geometric properties of objects. For example, while concrete, grass and streets are easily traversable, tree barriers and logs are not. Furthermore, due to the varying density of LIDAR point clouds, which are commonly used for data collection in driving scenes, the process of assessing distant areas can be inaccurate, leading to erroneous predictions. Our approach seeks to overcome these challenges to enable efficient and accurate real-time navigation in complex environments.

In this paper, we present FASTC, a novel deep learning-based method designed to overcome the aforementioned challenges by directly inferring the traversable classes from LIDAR scans. Our proposed approach includes three key components, which aim to leverage the strengths of deep learning to improve the accuracy of the traversability analysis. The first component, the pillar feature extraction module utilizes PointNet\cite{QiYi-18} to extract point features and 2D convolutions to extract spatial features and generate a high-dimensional feature map. The second one is the multi-frame fusion module that aggregates arbitrary multi-frame information through an attention module for a more accurate perception of distant areas. Finally, the traversability completion module completes the classification of traversability by learning the feature map to fill in the empty and recover the missing information, ultimately projecting the 3D data into a 2D bird's eye view (BEV) map. With these three modules, our FASTC can effectively and efficiently infer the semantic traversability classes of the environment and locate obstacles in real time for navigation. The key contributions of this work are summarized as follows: 
\begin{itemize}
  \item[1.]We propose a novel pillar feature extraction module and traversability completion module that operate directly on sparse 3D points and generate the traversability map. This results in a reduced computational cost while achieving better performance.
  \item[2.]We propose a spatio-temporal attention module to fuse multi-frame information, which effectively handles the varying density problem of LIDAR point clouds. This enables our module to accurately assess distant areas.
  \item[3.]Experimental results on the SemanticKITTI dataset \cite{behley2019SemanticKITTI} and the RELLIS-3D dataset \cite{jiang2021rellis} demonstrate the effectiveness of our method by outperforming strong baselines in both on-road and off-road settings.
\end{itemize}

 %------------------------------ Figure ---------------------------------------
\begin{figure*}[!htp]
\centering
\includegraphics[width=0.9\linewidth]{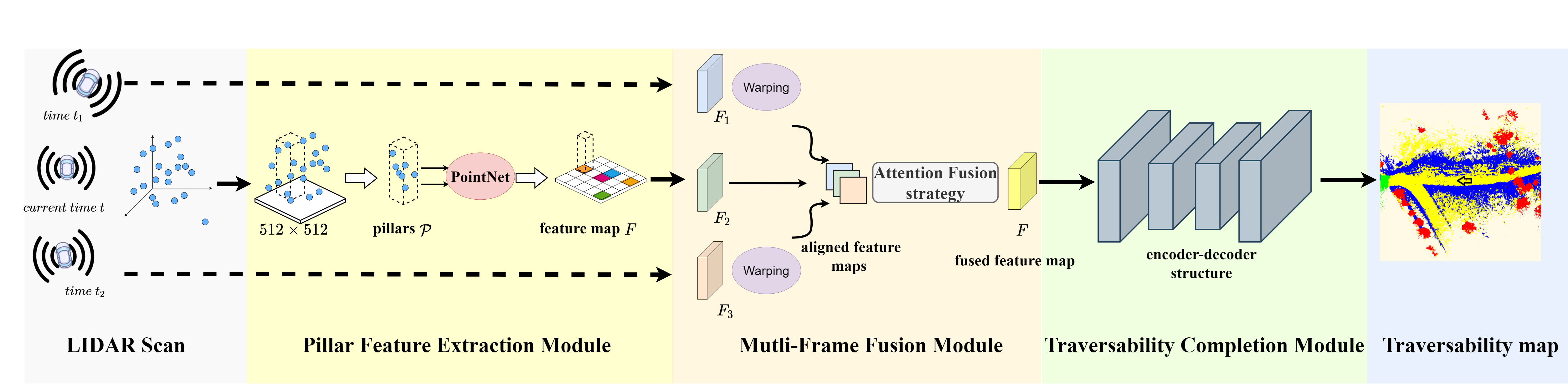} \\
\caption{\textbf{Overall architecture of FASTC. } FASTC  generates a semantic traversability map of the surrounding environment from the 3D LIDAR point cloud data in real-time. The framework consists of three main modules: (1) The pillar feature extraction module discretizes the raw point cloud into 2D grid-building pillars and uses PointNet to learn pillar-wise features, which are then converted into pseudo-images as feature maps. (2) The multi-frame fusion module aligns feature maps from multi frame scans captured at different times with the current frame's coordinate system and aggregates them using a fusion strategy based on a novel attention model. (3) The traversability completion module employs an encoder-decoder structure on the feature map F to fill in blank regions and generate the final BEV traversability map. }
\label{fig:kuangjia}
\end{figure*}
 %---------------------------------------------------------------------

\section{Related Work}
\textbf{LIDAR-based Semantic Segmentation.}  Based on the type of data input to neural networks, 3D point cloud segmentation methods can be classified into three categories\cite{GuoWang-7}: voxel-based, point-based, and view-based methods. Point-based methods\cite{QiSu-16,QiYi-18,ZhaoJiang-24} operate directly on raw point clouds and learn features from local points. Voxel-based methods\cite{ZhouZhu-55,choy20194d} convert point clouds into a voxel grid representation, where each voxel represents a small volume of space and sparse convolutions are then applied to the voxel grid. View-based point segmentation methods transform point clouds into various views or perspectives, such as range views\cite{51,milioto2019rangenet++} or BEV\cite{ZouLi-71,ZhangZhou-65}, to extract features and identify objects or regions in the data. These approaches allow the usage of  2D network architectures, which can be more efficient than 3D methods while still achieving high segmentation accuracy. Our method follows these research approaches and directly extracts point cloud features, projecting them onto a plane for segmentation.

\textbf{LIDAR-based Object Detection.} 3D object detection frameworks can be roughly classified into one-stage and two-stage methods. VoxelNet\cite{ZhouTuzel-58} is an end-to-end trainable framework that partitions a point cloud into equally spaced voxels and encodes voxel features into a 4D tensor, which is then processed by a region proposal network (RPN) \cite{ren2015faster} to obtain proposals. Improving upon VoxelNet, SECOND\cite{yan2018second} has achieved better performance and faster processing speed. LidarMultiNet \cite{ye2022LIDARmultinet} unifies semantic segmentation, panoptic segmentation, and object detection tasks, incorporating various types of supervision. Our approach is based on PointPillars\cite{lang2019pointpillars}, which efficiently organizes point clouds into vertical columns (pillars) and learns features from them. By extracting features directly from point clouds and encoding them into a standardized format, our method avoids the computationally expensive 3D convolution layers commonly used in object detection.

\textbf{Traversability Analysis.} Traversability refers to the suitability of terrain for driving based on physical properties such as slope, roughness, and surface condition. Identifying traversable regions is a significant criterion in autonomous driving\cite{suger2015traversability}, where on-road scenes are easily identifiable, but off-road scenes are challenging due to variations in terrain and the absence of visual identifiers. To estimate traversability from perception data, prior literature proposed several approaches, including manually designed methods \cite{fan2021step, wermelinger2016navigation}, conventional machine learning-based methods \cite{tang2019autonomous, belter2016adaptive}, and deep learning-based methods \cite{chavez2018learning}. Among these, learning-based approaches provide more flexibility and can be extended to different scenarios\cite{gao2021fine,shaban2022semantic}. Traversability information is commonly represented in a BEV (bird's eye view) map. However, prior literature \cite{sadat2020perceive, zeng2019end} rely on high-definition maps that are assumed to be available beforehand, which can be costly to produce. BEVNet \cite{shaban2022semantic} is a deep learning model that classifies terrain traversability in a local region around a mobile robot. However, we propose several new novel modules that help to achieve much better performance and at an even faster speed.   

\textbf{Attention Mechanism.}
Channel-wise attention was first used in the SENet\cite{hu2018squeeze} by utilizing global Average-pooling and fully connected layers to exploit inter-channel relationships. The Non-local networks\cite{wang2018non} adopt self-attention for video comprehension and object detection with the remarkable performance achieved. Convolutional block attention modules (CBAM)\cite{woo2018cbam} stack both channel attention and spatial attention in series. Li et al.\cite{li2020multigrained} utilized the attention module to focus on discriminative regions for fusing infrared and visible images, while UFA-Fuse\cite{zang2021ufa} introduced a novel and effective image fusion strategy based on unity fusion attention. Inspired by the work above, we also propose a spatial temporal attention module to fuse multi-frame features. This enables our model to aggregate information from multiple LIDAR scans, which can lead to a more accurate perception of the environment, especially for distant objects, and better production of the traversability map.

Our proposed method takes raw point clouds as input and generates a semantic traversability BEV  map with LIDAR location as the origin. The traversability map is classified into four levels: free, low-cost, medium-cost and lethal. The overview of the framework is illustrated in Fig \ref{fig:kuangjia}. It is composed of three main modules: pillar feature extraction, multi-frame fusion, and traversability completion. The pillar feature extraction module discretizes the input LIDAR scans into a 2D grid and constructs pillar-level features, which are then converted to feature maps in the form of pseudo-images. The multi-frame fusion module aligns the feature maps of multiple frames and fuses them using a novel attention model. The traversability completion module processes the feature maps, fills in the empty areas and produces the final traversability map. 

\subsection{Pillar Feature Extraction Module}
Unlike prior works\cite{shaban2022semantic,ZhouTuzel-58} that rely on computationally expensive 3D convolutions, the pillar feature extraction module divides the sparse point cloud according to $x-y$ coordinates to create pillars and casts sparse point cloud features in each pillar to generate a feature map.

\textbf{Point Cloud Discretization and Embedding.}
To leverage the point features, we discretize them into an evenly spaced grid in the x-y plane and organize them into vertical columns, creating a set of pillars $\mathcal{P} \text { with }|\mathcal{P}|=B$, similar to PointPillars \cite{LangVora-21}. We denote by $\mathbf{l}$ a point in the pillar with coordinates $x$, $y$, $z$ and reflectance $r$. The points in each pillar are then augmented with  $x_{c}$, $y_{c}$, $z_{c}$, $x_{p}$  and  $y_{p}$  where   $c$   denotes the distance to the arithmetic mean of all points in the pillar and the  $p$  denotes the offset from the pillar  $x, y$  center. The original point cloud $\mathbf{l}_{i}=\left[x_{i}, y_{i}, z_{i}, r_{i}\right]^{T} \in \mathbb{R}^{4}$ is then augmented to a $D$-dimension one with $D=9$, i.e.  $\mathbf{l}_{i_{aug}}=\left[x_{i}, y_{i}, z_{i}, r_{i}, x_{c}, y_{c}, z_{c}, x_{p}, y_{p}\right]^{T} \in \mathbb{R}^{9}$.

Due to the sparsity of the point cloud, the pillar set usually contains a large number of empty elements, and those that are not empty will typically contain only a small number of points. To address this, we fix the number of points ( $N$ ) per pillar and pre-define the number of pillars ( $P$ ) to create a dense input tensor of fixed size $(P, N, D)$. In cases where the number of pillars or LIDAR points does not reach the predefined value, we employ zero padding to generate a fixed-size pseudo-image. Conversely, if the number of pillars or points exceeds the desired value, we employ random sampling to ensure a fixed data structure.

\textbf{Pillar Feature Encoding. }
For each non-empty pillar of size  $(P, N, D)$, we employ a simplified version of PointNet\cite{QiSu-16} to learn the feature representation of every point within it. Our simplified PointNet consists of a linear layer, BatchNorm \cite{ioffe2015batch}, and ReLU\cite{nair2010rectified}, which then generates a tensor of size $(P,N,C)$. This tensor is then fed to a Max-pooling layer to produce an output tensor of size $(C, P)$. Finally, we place all pillar-wise features to their respective original locations in the grid, creating a  BEV representation of size $(C, H, W)$ where  $H$ and $W$ indicate the height and width of the grid. The generated feature map effectively encodes both the geometry and features of the point cloud scan in the form of pseudo-image.

\subsection{Multi-Frame Fusion Module}
A drawback of a LIDAR scan is that its point density varies with respect to the distance, whereas regions closer to the scanner have a higher point density than more distant areas. This can result in inaccurate traversability classifications especially for remote regions. To address this issue, we propose a fusion framework that leverages spatio-temporal information from multiple frames to improve the perception of sparse regions and mitigate the problem mentioned above.  This module first warps feature maps from multiple input scans, which are captured at different times, to align with the current frame's coordinate system. The aligned feature maps are then aggregated using a fusion strategy based on a novel attention model. 

%------------------------------ Figure ---------------------------------------
\begin{figure*}[!tp]
  \centering
  \includegraphics[width=\linewidth]{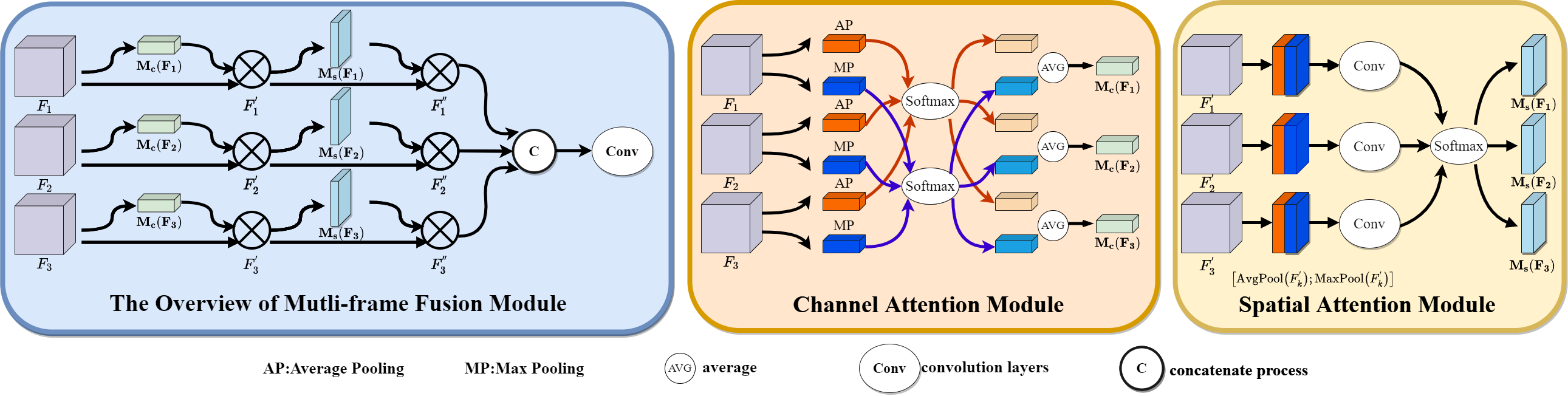} 
  \caption{\textbf{Overall architecture of our fusion module. } Our fusion module aims to produce a fused feature map from $K$ aligned intermediate feature maps. It consists of two components: a channel attention module and a spatial attention module. The channel attention module utilizes Max-pooling and Average-pooling with Softmax to identify significant information within each channel. Meanwhile, the spatial attention module uses the same pooling techniques to identify the relevant regions across all channels.}
 \label{fig:attention}
\end{figure*}
%---------------------------------------------------------------------
 
\textbf{Differentiable Warping.}	 
We address the issue of spatial non-synchronization among multiple frames by aligning different frames to the same odometry using a warping process. This process transforms the feature maps to the current odometry frame.

Specifically, given an intermediate feature map $F_{i,t_i}$ obtained from frame $t_i$ with transformation matrix $M_{t_i}$ and current frame transformation matrix $M_{t}$, we determine the relative transform of the odometry frame $t_i$ with respect to the current frame $t$. The transformation matrix $\mathcal{H}_{i}$ can be calculated as: $\mathcal{H}_{i}=M_{t_{i}}M_{t}^{-1}$,  which represents the transformation matrix from the odometry frame $t_i$ to the current frame $t$. To simplify the problem, we assume that there is no change in the LIDAR's z-axis. Under this assumption, the simplified matrix $\mathcal{H}_{i}$ can be represented as a 2x3 matrix with elements $h_{ij}$.

The warped feature map, denoted as $F_{i}$, is obtained by applying the transformation matrix $\mathcal{H}_i$ to the input feature map $F_{i,t_i}$. Mathematically, the warped feature map is expressed as:
\begin{eqnarray}
F_{i}(u, v)=F_{i,t_i}\left(\mathcal{H}_{i}[u, v, 1]^{T}\right)
\end{eqnarray}
where $(u,v)$ represents the spatial coordinate of the pixel. We further employ bilinear interpolation and zeros padding to get the final aligned features.
	
\textbf{Fusion Module.} 
Recent works utilize Recurrent Neural Network (RNN) to aggregate multi-frame information. However, This design is strongly influenced by the current input and may not effectively capture key information across multiple frames. We present a novel fusion approach that leverages attention mechanisms to effectively aggregate information from multiple frames.   Our fusion module is specifically designed to identify and enhance significant features through channel and spatial attention maps, ultimately producing the fused feature maps. The above procedure is illustrated in Fig. \ref{fig:attention}.

The proposed fusion module takes in $(K)$ intermediate feature maps, denoted as $F_{k} \in \mathbb{R}^{C \times H \times W}, \ k \in 1, \cdots, K$, where $K=3$ represents the number of frames of the input feature maps. The fusion module computes the channel and spatial attention maps, $M_{k}^{c} \in \mathbb{R}^{C \times 1 \times 1}$ and $M_{k}^{s} \in \mathbb{R}^{1 \times H \times W}$, based on the input feature maps. The feature maps are then multiplied by these channel and spatial attention maps to obtain the attention-weighted feature maps. Finally, the attention-weighted feature maps are concatenated and fed into a convolution of kernel size $1\times1$ to reduce the dimension and generate the fused feature map $F\in \mathbb{R}^{C \times H \times W}$. The complete process can be described as: 
%-----------------------------------------------------------
\begin{eqnarray}
\centering
\begin{gathered}
F_{k}^{\prime}=M_{c}\left(F_{k}\right)\otimes F_{k}\\
F_{k}^{\prime\prime}=M_{s}\, (F_{k}^{\prime})\otimes F_{k}^{\prime}\\
F=f^{1 \times 1}\left([F_1^{\prime\prime};F_2^{\prime\prime};\cdot\cdots  ;F_k^{\prime\prime}]\right)
\end{gathered}
\end{eqnarray}
%--------------------------------------------------------

where the $\otimes$ denotes element-wise multiplication, 
$F_{k}^{\prime}$ and $F_{k}^{\prime\prime}$  denote the intermediate outputs of the fusion module. The following describes the details of each module.

(1) Channel attention module:  The channel attention map $M_{c}(F_k)$ for the feature map $F_k$ is generated through the following procedure:
%-------------------------------------------------------
\begin{eqnarray} 
\begin{gathered}
{AF}_k  ={\operatorname{AvgPool}(F_k)},\ {MF}_k  ={\operatorname{MaxPool}(F_k)}\\
M_{c}(F_k) = \frac{1}{2} \Big( \frac{\exp(AF_k)}{\sum_{j=1}^K \exp(AF_{j})} +\frac{\exp(MF_k)}{\sum_{j=1}^K \exp(MF_{j})}\Big)
\end{gathered}
\end{eqnarray}
%-------------------------------------------------------------------
where ${AF}_k \in \mathbb{R}^{C \times 1 \times 1}$ and ${MF}_k \in \mathbb{R}^{C \times 1 \times 1}$ represent the channel Average-pooling and Max-pooling features. Subsequently, the Softmax activation function is applied to these channel features to generate the channel attention map.  The channel attention exploits the feature relationship of multi-frame feature maps and highlights the important channel information. This provides crucial information on "what" to fuse.

(2) Spatial attention module: The spatial attention map $M_{s}(F^{\prime}_k)$ for the channel-attention feature map $F^{\prime}_k$  is  generated through the following procedure:
%---------------------------------------------------------------------
\begin{eqnarray} 
\begin{gathered}
M_{k}= f^{7 \times 7}\Big(\left[\operatorname{AvgPool}\left(F_{k}^{\prime}\right) ; \operatorname{MaxPool}\left(F_{k}^{\prime}\right)\right]\Big) \\
M_{s}(F_k)= \frac{\exp({M}_{k})}{\sum_{i=1}^{K} \exp({M}_{k})}
\end{gathered}
\end{eqnarray}
%---------------------------------------------------------------------
where $\operatorname{AvgPool}\left(F_{k}^{\prime}\right) \in \mathbb{R}^{1 \times H \times W}$ and  $\operatorname{MaxPool}\left(F_{k}^{\prime}\right) \in \mathbb{R}^{1 \times H \times W}$ represent the spatial Average-pooling and Max-pooling features. $f^{7 \times 7}$ denotes the $7\times7$ convolution operation, operates on 
the concatenated spatial features to obtain spatial attention features ${M}_{k}\in \mathbb{R}^{1 \times H \times W}$. Then the  Softmax activation function is applied to these spatial features to generate the spatial attention map. Spatial attention captures the spatial information and identifies significant regions. This provides crucial information on "where" to fuse.

\subsection{Traversability Completion Module}
The traversability completion module adopts an encoder-decoder structure on the previously obtained feature map $F$ to capture local information with a large field-of-view for filling in blank regions and generating the final BEV traversability map.  This module is inspired by the network structure of Regseg\cite{gao2021rethink}.

%------------------------------ Figure ---------------------------------------
\begin{figure}[!tp]
  \centering
  \includegraphics[width=\linewidth]{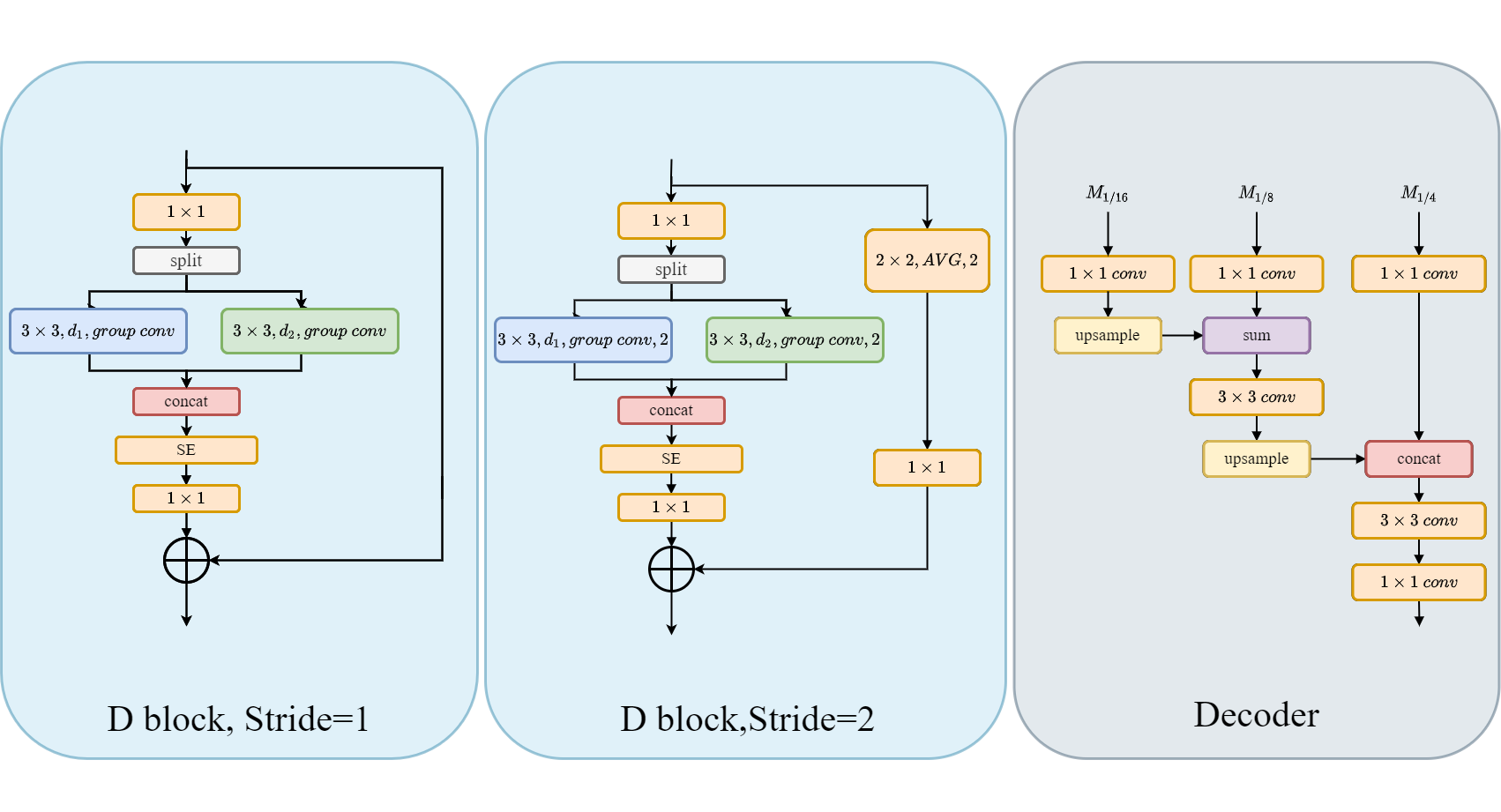} 
  \caption{\textbf{Dilated block and lightweight decode.} Our D block utilizes group convolutions and the lightweight decoder efficiently generates the traversability map.
  }
 \label{fig:Dblock_and_decoder}
\end{figure}
%--------------------------------------------------------------------

%------------------------------ Table ---------------------------------------
\begin{table}[!t]
\centering
\begin{tabular}{ccccc}
\hline
              & (d1,d2) & stride & output channel & repeat \\ \hline
Dilated block & 1,1     & 2    & 64             & 1      \\
Dilated block & 1,1     & 2    & 96             & 1      \\
Dilated block & 1,1     & 1    & 128            & 2      \\
Dilated block & 1,1     & 2    & 256            & 1      \\
Dilated block & 1,1     & 1    & 256            & 1      \\
Dilated block & 1,2     & 1    & 256            & 1      \\
Dilated block & 1,4     & 1    & 256            & 4      \\
Dilated block & 1,14    & 1    & 256            & 6      \\
Dilated block & 1,14    & 1    & 320            & 1      \\ \hline
\end{tabular}
\caption{\textbf{Backbone of encoder. } Channels is the number of output channels, and the number of input channels is inferred from the previous block. When stride = 2 and repeat > 1, the first block has stride 2 and the rest have stride 1.}
\label{encoder}
\end{table}
%---------------------------------------------------------------------

\textbf{Dilated block.} Our network adopts dilated blocks as the fundamental elements same as \cite{gao2021rethink}. These blocks utilize group convolutions with different dilation rates across groups to extract multi-scale features. The overall architecture of our D block with stride = 1 and 2 are illustrated on the left part of Fig.\ref{fig:Dblock_and_decoder}, where ($d_1$, $d_2$) denotes the dilation rates of the dilated blocks. The BatchNorm and ReLU follow each convolution and we utilize the Squeeze-and-Excitation Network\cite{hu2018squeeze} reduction ratio of $\frac{1}{4}$.

\textbf{Encoder.} The encoder module consists of various dilated blocks with different ($d_1$, $d_2$) settings. We use ($d_1$, $d_2$) to represent the structure of each block and provide the detailed architectures in Tab.~\ref{encoder}. Specifically, the module starts with a 64-channel dilated block and a 96-channel dilated block produce the feature map $M_{1/4}$ at $\frac{1}{4}$ resolution, followed by three 128-channel dilated blocks that generate feature map $M_{1/8}$ at $\frac{1}{8}$ resolution, and thirteen 256-channel dilated blocks, with one final 320-channel dilated block producing feature map $M_{1/16}$ at $\frac{1}{16}$ resolution.

\textbf{Decoder.}  The decoder captures the features and generates the final traversability map. Each resolution feature map corresponds to a decoder layer. $1\times1$ convolutions with 128 output channels are applied to $M_{1/16}$ and $M_{1/8}$ and $1\times1$ convolutions with 8 output channels is applied to $M_{1/4}$. The up-sampling operation is performed on the $M_{1/16}$ branch and added to the $M_{1/8}$ branch to produce $M_{1/8}'$. Subsequently, $M_{1/8}'$ is fed into $3\times3$ convolutions with 64 output channels, followed by an up-sampling operation and concatenation with the $M_{1/4}$ feature map. The consolidated feature map $M_{1/4}'$ is fed into $3\times3$ convolutions with 64 output channels and $1\times1$ convolutions with 5 output channels. BatchNorm and ReLU are applied to all convolutions except for the final one. The detailed decoder architecture is illustrated on the right part of Fig.~\ref{fig:Dblock_and_decoder}.

\section{Implementation Details}
\subsection{Loss Function and Evaluation Metrics}
Our method takes LIDAR scans as input and outputs the corresponding traversability map. We train the network in a supervised manner, where we use the cross entropy loss between the ground-truth label and the predicted label at each pixel as the loss function. We find that this loss function is sufficient to achieve promising results and is easy to optimize.

To evaluate the quality of our predictions and inference speed, we use mean intersection over union (mIoU) and mean accuracy (mAcc) as quantitative measures of prediction accuracy. We compute mIoU and mAcc as follows: $mIoU=\frac{1}{C} \sum_{i=1}^{C} \frac{TP_{i}}{TP_{i}+FP_{i}+FN_{i}}$, $mAcc=\frac{1}{C} \sum_{i=1}^{C} \frac{TP_{i}}{TP_{i}+FP_{i}}$, where $C$ represents the total number of classes, $TP_{i}$, $FP_{i}$, and $FN_{i}$ represent true positive, false positive, and false negative predictions for class $i$. We also measure the speed as a quantitative evaluation metric. Note that the dataset includes an additional "unknown" class, which is only included in the training procedure but excluded during the evaluation step.

%------------------------------ Table---------------------------------------
\begin{table*}[!tp]
\centering
\begin{tabular}{c|c|ccccc}
\hline
\multirow{2}{*}{Method}                & \multirow{2}{*}{\textit{Settings}}  & \multicolumn{2}{c}{SemanticKITTI}                               & \multicolumn{2}{c}{RELLIS-3D}                                   & \multirow{2}{*}{speed(fps)↑}  \\
                                       &                                     & mIoU(\%)↑                      & \multicolumn{1}{c|}{mAcc(\%)↑} & mIoU(\%)↑                      & \multicolumn{1}{c|}{mAcc(\%)↑} &                               \\ \hline
BEVNet-S                               & \multirow{2}{*}{single frame input} & 41.6                           & \multicolumn{1}{c|}{59.9}      & 55.9                           & \multicolumn{1}{c|}{76.9}      & 10                            \\
\textbf{Ours FASTC-S} &                                     & \textbf{48.4} & \multicolumn{1}{c|}{66.7}      & \textbf{61.7} & \multicolumn{1}{c|}{82.1}      & \textbf{20}  \\ \hline
Cylinder3D-TA-3D       & \multirow{5}{*}{multi frame inputs} & 47.1                           & \multicolumn{1}{c|}{N/A}       & 40.8                           & \multicolumn{1}{c|}{N/A}       & N/A                           \\
Cylinder3D-TA                          &                                     & 46.5                           & \multicolumn{1}{c|}{N/A}       & 41.1                           & \multicolumn{1}{c|}{N/A}       & N/A                           \\
BEVNet-TA                              &                                     & 46.8                           & \multicolumn{1}{c|}{N/A}       & 61.5                           & \multicolumn{1}{c|}{N/A}       & N/A                           \\
BEVNet-R                               &                                     & 53.5                           & \multicolumn{1}{c|}{69.16}     & 64.4                           & \multicolumn{1}{c|}{79.2}      & \textbf{6.3} \\
\textbf{Ours FASTC-M} &                                     & \textbf{54.4} & \multicolumn{1}{c|}{69.38}     & \textbf{68.6} & \multicolumn{1}{c|}{82.7}      & 6.2                           \\ \hline
\end{tabular}
\caption{ \textbf{Quantitative results of  traversability classification.} We report the metrics of mIoU, mAcc, and speed for both SemanticKITTI and RELLIS-3D datasets. Our FASTC  outperforms other learning based methods and achieve high accuracy and speed.}
\label{tab1}
\end{table*}
%--------------------------------------------------------------------

\subsection{ Dataset Generation}\label{section:dataset}
Our method requires LIDAR scans as the input and the ground-truth traversability label maps for the loss function, We evaluate the performance of our method on both on-road and off-road scenes. To get the above-required data, we follow the principle of BEVNet\cite{shaban2022semantic} to generate the traversability datasets from SemanticKITTI \cite{behley2019SemanticKITTI} and RELLIS-3D \cite{jiang2021rellis}. SemanticKITTI provides point-wise annotation of point cloud sequences for on-road scenes, while RELLIS-3D provides point-wise annotations representative of off-road scenes.

Specifically, traversability datasets are generated through the mapping of semantic classes to traversability classes via the aggregation of $N$ scans with stride $S$, performing points binning and ground height estimation, and conducting traversability projection operations. The dataset is composed of the original LIDAR scan, pose, and a 2D BEV traversability grid map at each scan pose. The dataset classifies the surrounding environment into 4-level traversability classes (free, low-cost, medium-cost, and obstacle) and an additional unknown class to represent undiscovered regions. Our dataset is similar to BEVNet's original setting, with a value of $N$ = 71, $S$ = 2 for SemanticKITTI and $N$ = 141, $S$ = 5 for RELLIS-3D. The 2D traversability maps are sized into 51.2 $m$ × 51.2 $m$ with the resolution of 0.2 $m$.

%------------------------------ Figure ---------------------------------------
 \begin{figure}[!htp]
  \centering
  \includegraphics[width=8cm]{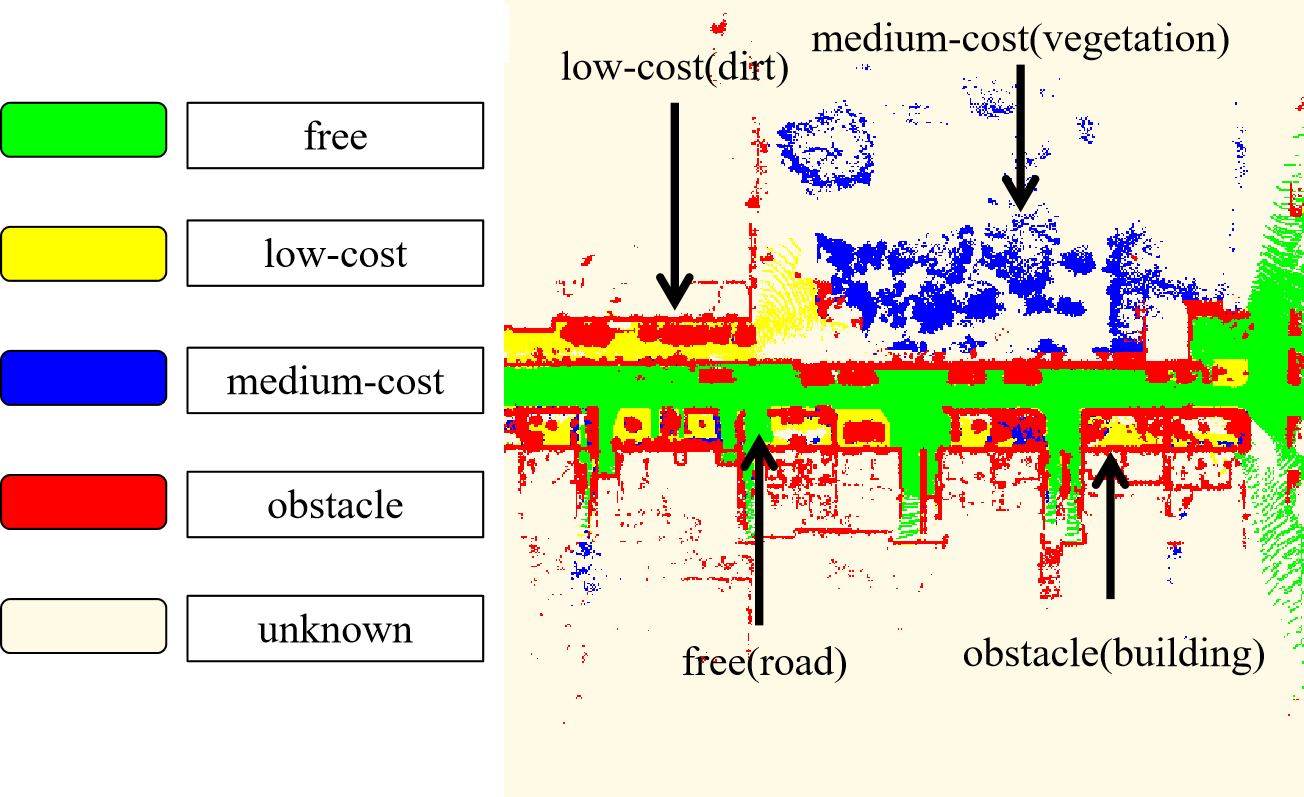} 
  \caption{\textbf{Example of Traversability Dataset. }The semantic traversability map of the surrounding environments indicates different traversability categories with distinct colors.}
 \label{fig:sample}
\end{figure}
%---------------------------------------------------------------------

\subsection{Training Details}
We adopt P = 80000, N = 55, C = 128, H = 512 and W = 512. The input point cloud is cropped at  [(-51.2, 51.2), (-51.2, 51.2), (-3, 3)] meters along x, y, and z axes respectively. When generating
pillars, we use a pillar grid size of $0.2^2 m^2$. For the training process, our work utilizes the Adam optimizer \cite{kingma2014adam} with an initial learning rate of 2.0e-4 and a weight decay of 0.01. Firstly, we train a single-frame model without a multi-frame fusion module until convergence. Then, we insert the multi-frame fusion module and prohibit gradient propagation to the pillar feature extraction module. We jointly train the multi-frame fusion module and the traversability completion module. This two-stage training procedure is fast and memory-efficient.

\section{Experiments}

We evaluate our method on both on-road and off-road scenes from the SemanticKITTI and RELLIS-3D datasets. We use BEVNet as our baseline and compare our method with different variants, including BEVNet-S (single-frame input), BEVNet-TA (temporal aggregation) and BEVNet-R (recurrent). Additionally, we compare our method against various models based on Cylinder3D \cite{ZhouZhu-55}, including Cylinder3D-TA (fine-tuned on our 4-class ontology with temporal aggregation) and Cylinder3D-TA-3D (using Octomap with temporal aggregation)\footnote{The quantitative results of  Cylinder3D-TA and  Cylinder3D-TA-3D are directly obtained from the reported results in \cite{shaban2022semantic} }. Our proposed model can be divided into two categories: FASTC-S, which utilizes a single-frame scan as input, and FASTC-R, which uses multi-frame scans as input. The results of our study are presented in Tab.~\ref{tab1} for quantitative evaluation and in Tab.~\ref{tab2} for class IoU evaluation.

%------------------------------ Table ---------------------------------------
\begin{table}
\centering
\resizebox{\linewidth}{!}
{
\begin{tabular}{c|cccc}
\hline
\multirow{2}{*}{}     & \multicolumn{4}{c}{SemanticKITTI(\%)}                         \\
                      & free\textcolor{green}{$\blacksquare$ } & low-cost\textcolor{yellow}{$\blacksquare$ } & medium-cost\textcolor{blue}{$\blacksquare$ } & lethal\textcolor{red}{$\blacksquare$ } \\ \hline
BEVNet-S              & 64.06     & 48.12         & 15.01            & 39.21       \\
Ours FASTC-S              & 67.05     & 50.00         & 28.01            & 48.70       \\ \hline
BEVNet-R              & 76.80     & 62.00         & 21.50            & 53.82       \\
Ours FASTC-M             & 74.54     & 55.91         & 29.12            & 58.27       \\ \hline
\multicolumn{1}{l|}{} & \multicolumn{4}{c}{RELLIS-3D(\%)}                              \\ \hline
BEVNet-S              & 65.30     & 54.17         & 52.21            & 51.6       \\
Ours FASTC-S             & 66.10     & 64.61         & 55.13            & 60.96       \\ \hline
BEVNet-R              & 67.50     & 59.30         & 63.30            & 67.60       \\
Ours FASTC-M             & 75.45     & 66.70         & 63.65            & 68.30       \\ \hline
\end{tabular}
}
\caption{ \textbf{Detailed comparison result on traversability classes.} We show IoU of each traversability class on SemanticKITTI and RELLIS-3D.}
\label{tab2}
\end{table}
%---------------------------------------------------------------------

\subsection{Comparisons on Results with Single Frame Input.} 
We evaluate our approach for traversability classification tasks with single-frame input. As shown in Tab.~\ref{tab1}, Our approach achieves high accuracy and operates at a fast speed of 20 fps, outperforming existing works. Visual comparisons of our method FASTC-S with BEVNet-S are presented in Fig. \ref{tab2}.

\begin{figure}[!t]
\centering
\setlength{\tabcolsep}{0.8mm}
\begin{tabular}{ccc}
\includegraphics[width=0.31\linewidth]{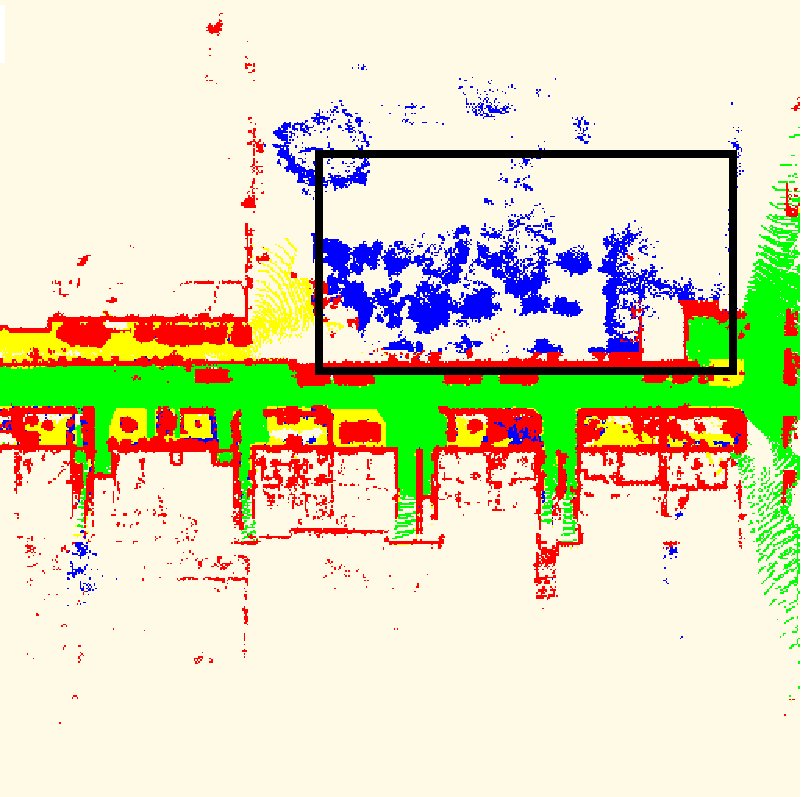}&
\includegraphics[width=0.31\linewidth]{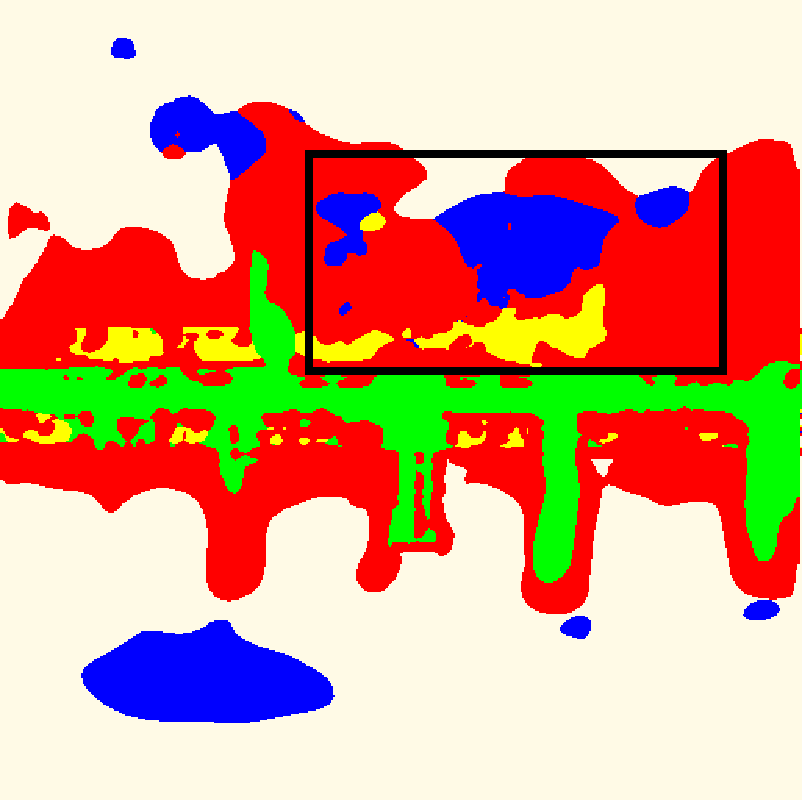}&
\includegraphics[width=0.31\linewidth]{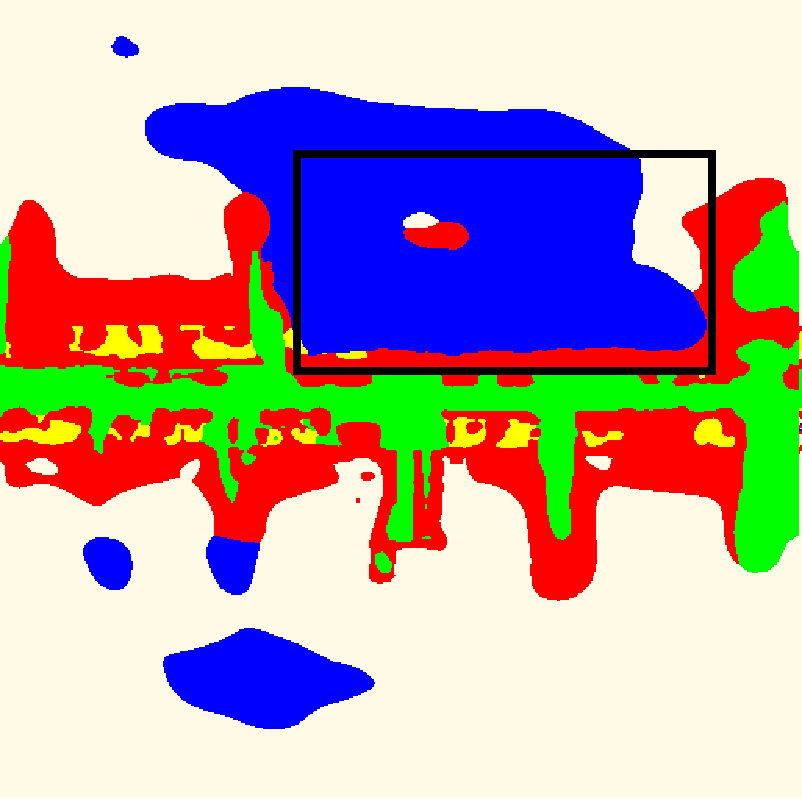}\\
\includegraphics[width=0.31\linewidth]{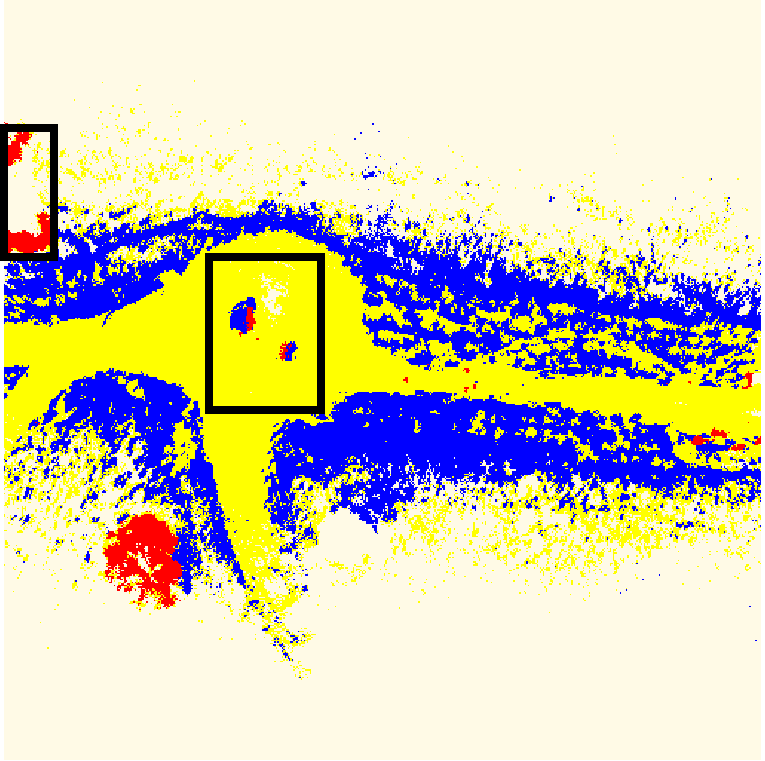}&
\includegraphics[width=0.31\linewidth]{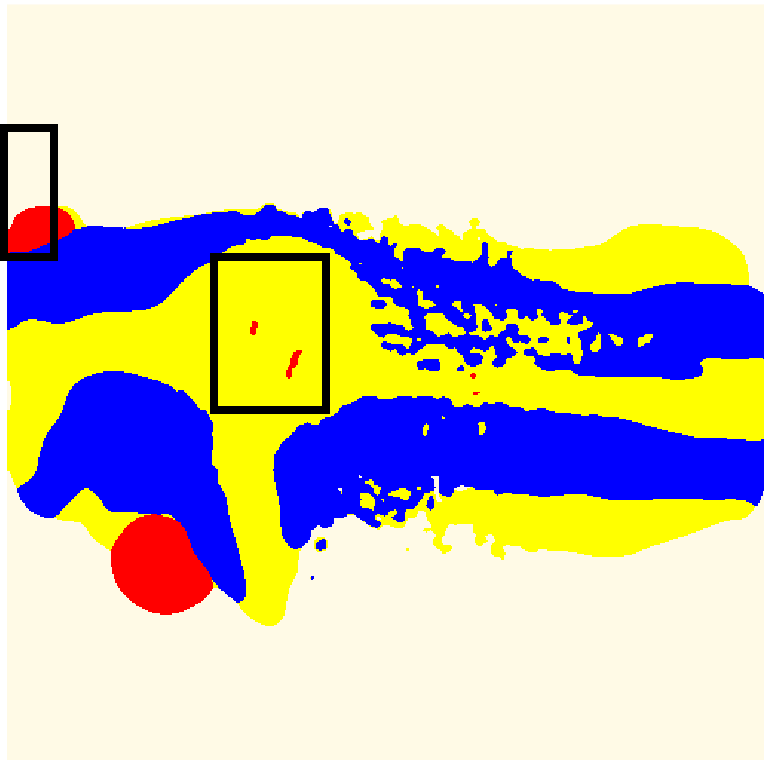}&
\includegraphics[width=0.31\linewidth]{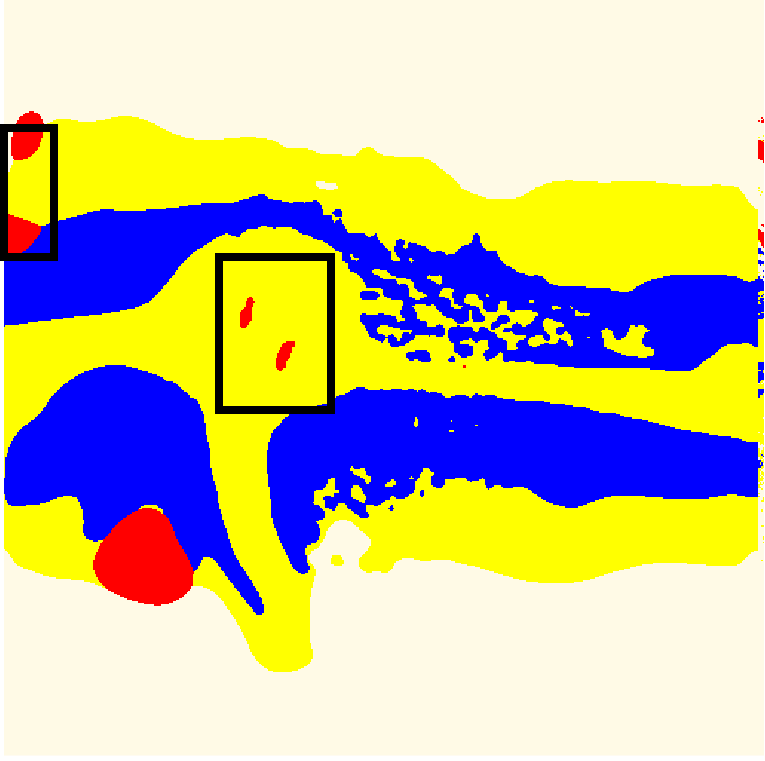}\\
(a) Ground Truth & (b) BEVNet-S & (c) FASTC-S
\end{tabular}
\caption{\textbf{Qualitative comparison of single input on SemanticKITTI(top) and RELLIS-3D(bottom).} Our FASTC-S performs better in catching the local information to fill the empty (top) and distinguishing the obstacles with superior obstacle locating (bottom). } 
\label{fig:single}
\end{figure}

\textbf{Results on SemanticKITTI.}  
Our method achieves higher precision in terms of mIoU and mAcc on on-road scenes, as shown in Tab.~\ref{tab1} and Tab.~\ref{tab2}. Specifically, it demonstrates an improved ability to distinguish medium-cost and lethal classes, with 13.00\% and 9.49\% improvements. Our method captures the regular road structure of the SemanticKITTI dataset, filling in empty and unknown regions, as shown in Fig. \ref{fig:single}. In contrast to BEVNet, which produces false results, our method more accurately distinguishes the distribution of obstacles and locates them.

\textbf{Results on RELLIS-3D.} 
Our method outperforms the compared method on off-road scenes especially in the lethal and low-cost class, with improvements of 9.36\% and 10.34\%,  as demonstrated in Tab.~\ref{tab1} and Tab.~\ref{tab2}.  As shown in Fig. \ref{fig:single}, our method accurately identifies obstacles and produces a more accurate road profile. The off-road environment presents more challenges and makes it difficult for 3D segmentation methods due to the lack of environmental structure. However, our FASTC-S model can effectively handle the complex environment, process the data, and produce reliable traversability classification results.

\subsection{Comparisons on Results with Multi Frames Input.}\label{section:mulit fuse}
\begin{figure}[!t]
\centering
\setlength{\tabcolsep}{0.5mm}
\begin{tabular}{cccc}
\includegraphics[width=0.24\linewidth]{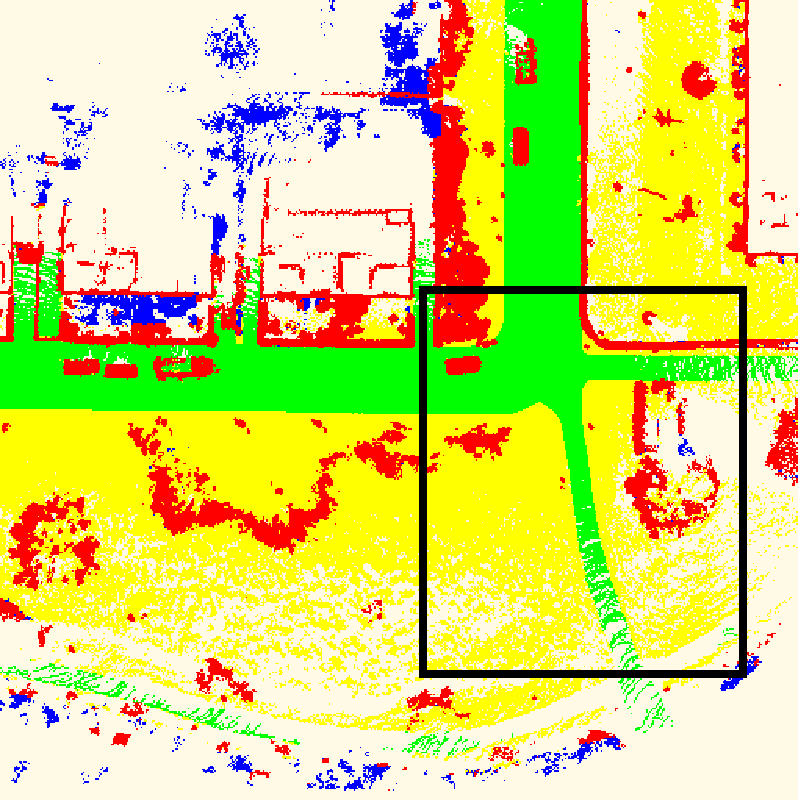}&
\includegraphics[width=0.24\linewidth]{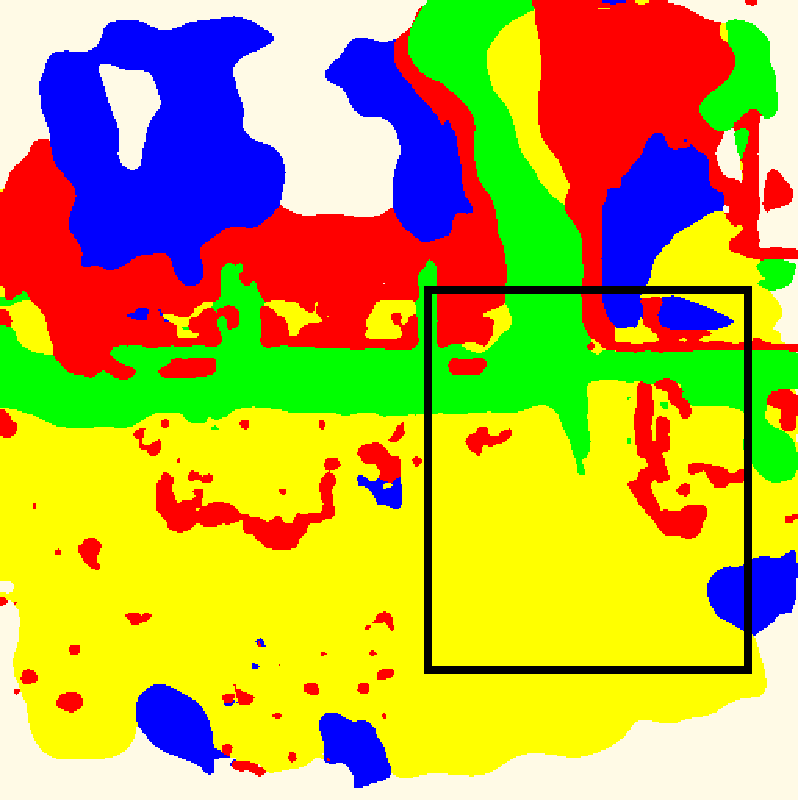}&
\includegraphics[width=0.24\linewidth]{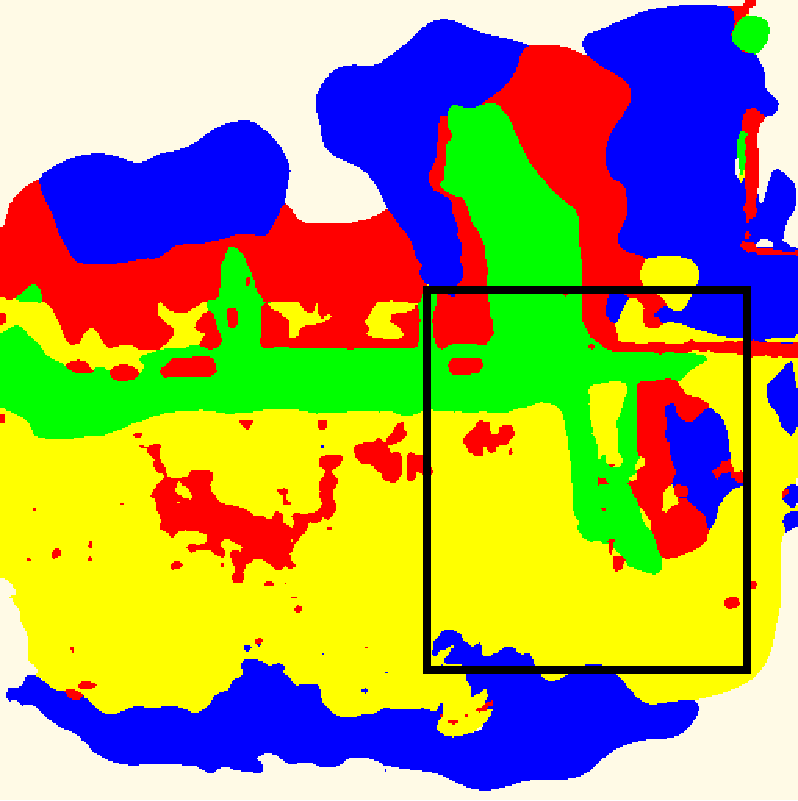}&
\includegraphics[width=0.24\linewidth]{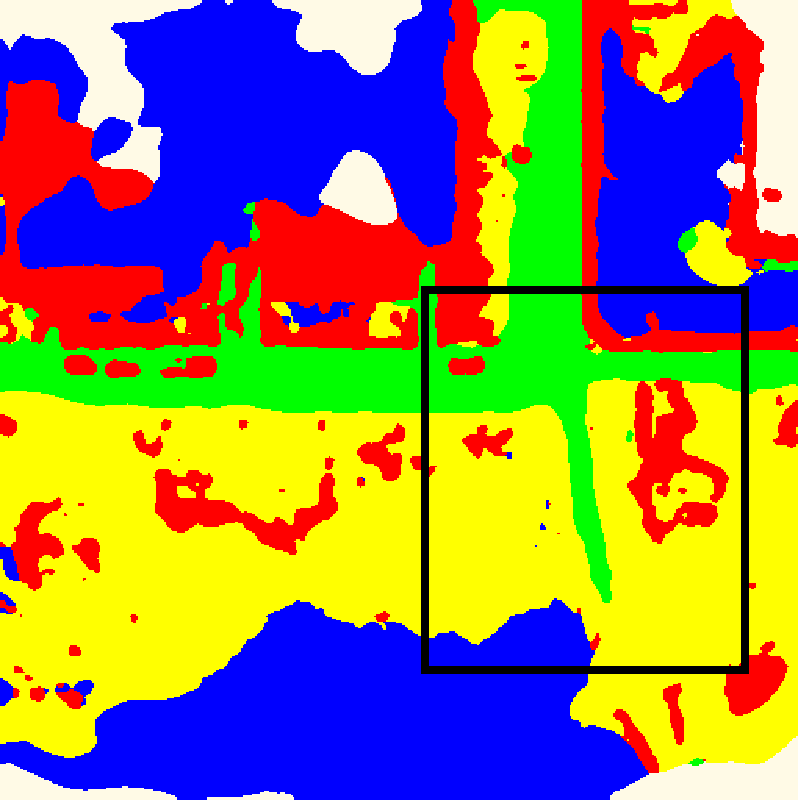}\\
\includegraphics[width=0.24\linewidth]{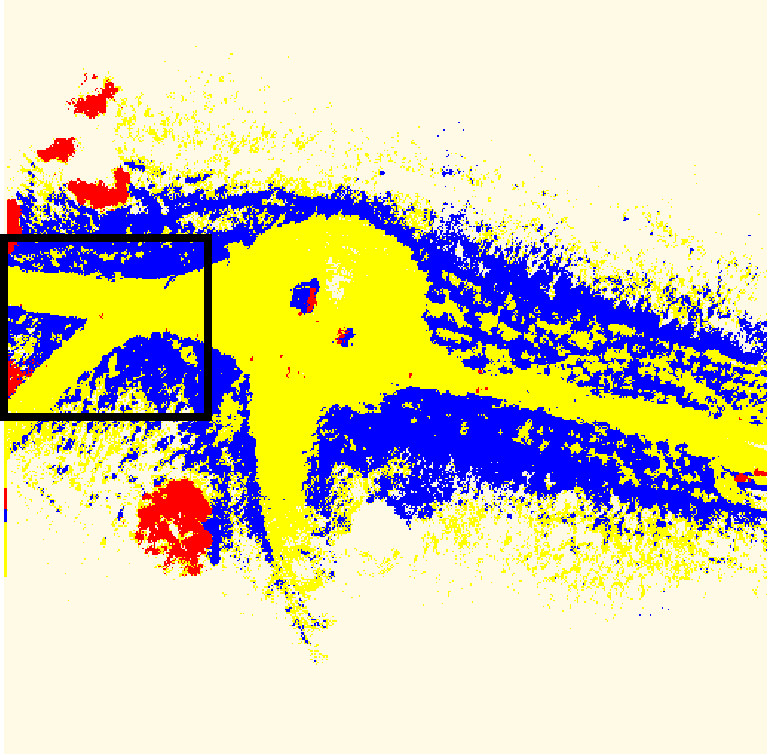}&
\includegraphics[width=0.24\linewidth]{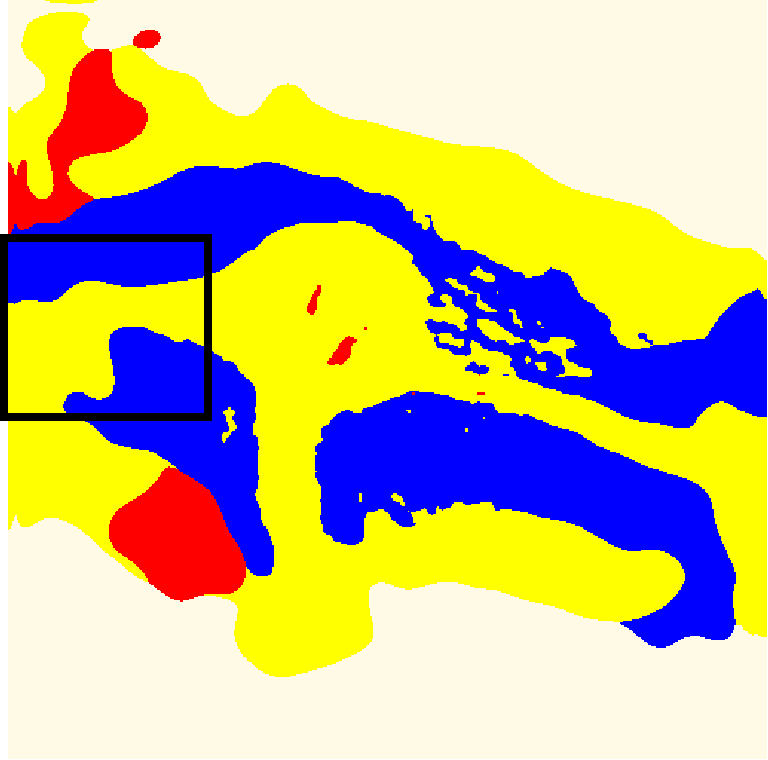}&
\includegraphics[width=0.24\linewidth]{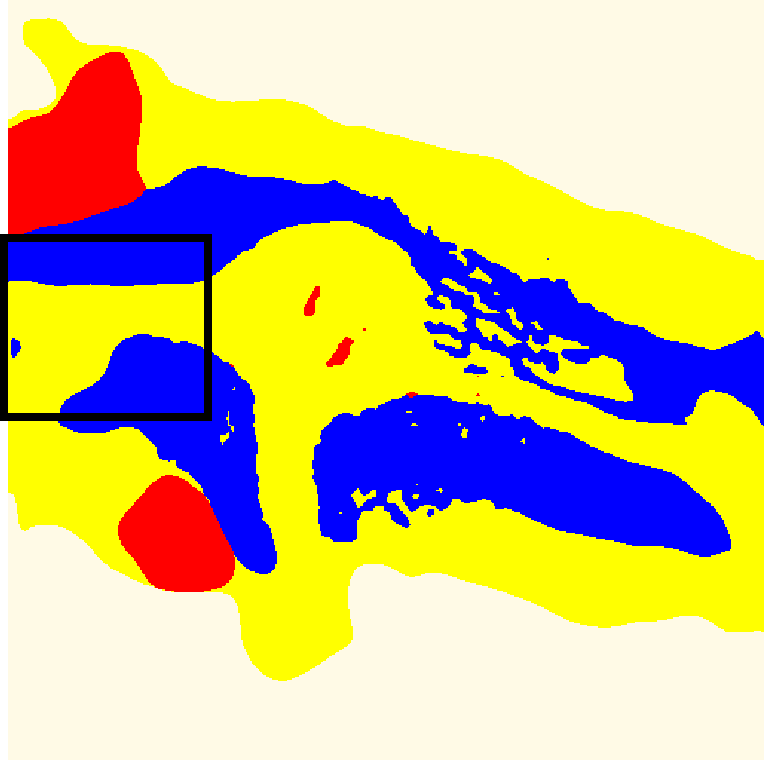}&
\includegraphics[width=0.24\linewidth]{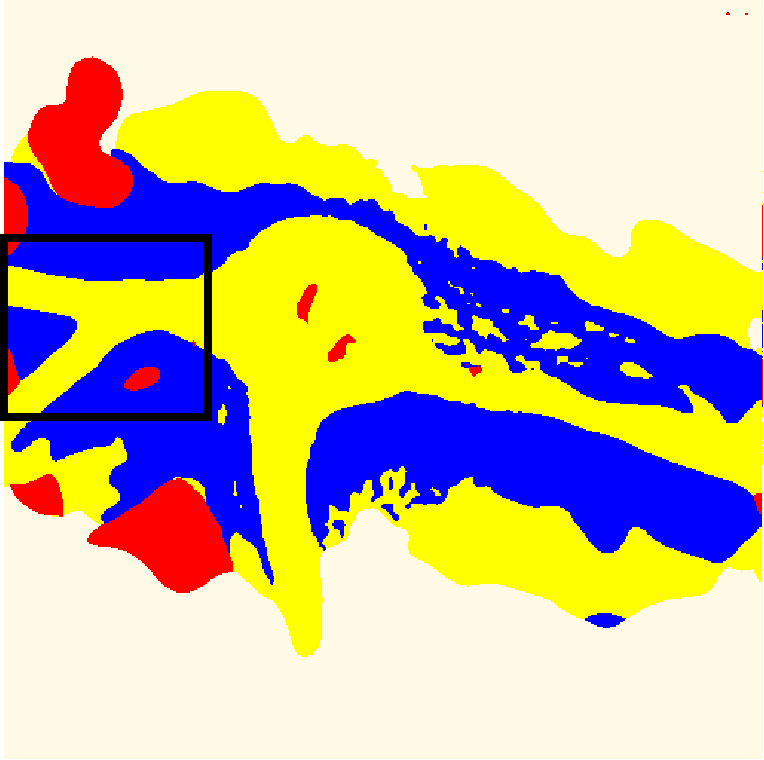}\\
(a)\footnotesize{Ground\,Truth} & (b) \footnotesize{FASTC-S} & (c) \footnotesize{BEVNet-R} & (d) \footnotesize{FASTC-M}
\end{tabular}
\caption{\textbf{Qualitative comparison of single input on SemanticKITTI(top) and RELLIS-3D(bottom).} Our FASTC-M excels in identifying road trends and accurately locating small obstacles. } 
\label{fig:multi}
\end{figure}

We further evaluate our approach for traversability classification tasks with multi-frame inputs, which include two additional input scans. Tab.~\ref{tab1} shows that our approach effectively fuses multi-frame information to support inference and achieves a processing fast speed of 6.2 fps, comparable to BEVNet-R. Fig. \ref{fig:multi} illustrates the visual comparisons of our FASCT-F method with BEVNet-R and the single-scan method FASTC-S.

\textbf{Results on SemanticKITTI.} 
As shown in Tab.~\ref{tab1} and Tab.~\ref{tab2}, our FASTC-M outperforms BevNet-R by 1\% and FASTC-S by 7\% in mIoU, which demonstrates the effectiveness of our multi-frame fusion model on on-road scenes, especially in distinguishing the medium-cost class. FASTC-M effectively distinguishes remote regions and produces accurate inferences at a large scale and exhibits excellent performance in handling serious occlusions and complex clutter changes caused by moving vehicles and bikes. In contrast, BEVNet-R is incapable of handling different types of information and produces sparse results, while FASTC-S fails to provide accurate results in regions with complex clutter changes and occlusions. Through visual comparisons in Fig.~\ref{fig:multi}, our approach excels in identifying road trends and accurately locating potential obstacles on the roadside. Additionally, with the inclusion of a memory module, BEVNet-R outperforms FASTC-M in free and low-cost classes, exhibiting superior ability to remember the trajectory traversed in specific areas.

%------------------------------ Figure ---------------------------------------
%---------------------------------------------------------------------

\textbf{Results on RELLIS-3D.} FASTC-M, demonstrates superior performance in off-road scenes compared to BEVNet-R, with a 4\% improvement and a 7\% improvement in FASTC-S in mIoU as shown in Tab.~\ref{tab1}. FASTC-M performs significantly well on the free cost and low-cost classes as shown in Tab.~\ref{tab2} and effectively handles complex information. As illustrated in Fig.~\ref{fig:multi}, our approach excels in accurately predicting road trajectories, while also identifying small, independent obstacle regions and vegetation in the far distance.

\section{Ablation Study} 
We perform an ablation study to analyze the impact of various components in our proposed network architecture, FASTC, with the aim of gaining a deeper understanding of the underlying mechanisms. Specifically, we investigate different variants of our model and assess the effects of the proposed modules (discussed in Section \ref{section：module}), as well as the impact of the fusion strategy (discussed in Section \ref{section：order}).

\subsection{Effectiveness of Proposed Module} \label{section：module}
We conduct a comprehensive analysis of each component in our proposed approach, as shown in Tab.~\ref{table:abli}. BEVNet can be roughly described as a network architecture consisting of 3D Convolution, ConvGRU\cite{shi2015convolutional} and HardNet\cite{chao2019hardnet}. To identify the contribution of each module, we replaced them one by one and evaluated their effect.

 \textbf{Ablation for pillar feature extraction module.} As shown in Tab.~\ref{table:abli}, our proposed pillar feature extraction module demonstrates superior performance in both speed and accuracy compared to the 3D convolution method. The ablation study shows that the module effectively captures local point features and generates precise feature maps. Furthermore, its vertical column (pillar) structure enables significantly lower memory usage.  Our module improves performance by approximately 1\% and increases speed from 8fps to 20fps.

\textbf{Ablation for multi-frame fusion module. }Our ablation study demonstrates the efficacy of our multi-frame fusion module in capturing significant features from different frame feature maps, resulting in improved prediction accuracy. Notably, our module achieves this without relying on recurrent neural networks, leading to a performance increase of approximately 2\%, as shown in Tab.~\ref{table:abli}.

\textbf{Ablation for traversability completion module.}  Our ablation study reveals that the superior performance of our proposed traversability completion module is attributed to the utilization of dilated convolutions, which effectively enlarges the perceptual field of the network and captures more information. Our experiments on the SemanticKITTI and RELLIS-3D datasets achieve an improvement of 4\% in accuracy compared to the baseline, as shown in Tab.~\ref{table:abli}.
%------------------------------ Figure ---------------------------------------
%--------------------------------------------------------------------

%------------------------------ Table ---------------------------------------
\begin{table}[!t]
\centering
\resizebox{\linewidth}{!}{
\begin{tabular}{c|ccc}
\hline
\multirow{2}{*}{method}  & \textit{SemanKITTI} & RELLIS-3D & \multirow{2}{*}{Speed(fps)} \\
                         & \multicolumn{2}{c}{MIOU(\%)}    &                             \\ \hline
Pillar+HardNet           & 44.2                & 57.3      & 23                          \\
3D Conv+Completion           & 47.1                & 59.9      & 8                           \\
FASTC-S                  & 48.4                & 61.7      & 20                          \\ \hline
3D Conv+Attention+Completion & 53.9                & 67.3      & 3.9                         \\
Pillar+ConvGRU+Completion    & 52.5                & 66.5      & 12.2                        \\
Pillar+Attention+HardNet & 53.5                & 67.1      & 7.6                        \\
FAST-M                   & 54.4                & 68.6      & 6.2                         \\ \hline
\end{tabular}
}
\caption{ \textbf{Performance comparison of each component:} The pillar component denotes the Pillar Feature Extraction Module, Attention denotes the Multi-Frame Fusion Module and Completion denotes the Traversability Completion Module.}
\label{table:abli}
\end{table}
%---------------------------------------------------------------------

\subsection{Effectiveness of Order of Fusion Stragtegy} \label{section：order}
In our proposed approach for processing multi-frame LIDAR scans, we adopt a sequential pipeline comprising a wrapping function, a fusion module and a traversability completion module. We refer to the fusion strategy as pre-fusion. In this study, we explore two additional strategies for multi-frame fusion, namely in-fusion, and post-fusion. The fusion operation is placed at the end of the pipeline to obtain the final output in both in-fusion and post-fusion methods. In the in-fusion strategy, the wrapping operation is performed before the traversability completion module, while in the post-fusion method, it is performed after.

Tab.~\ref{tab3} demonstrates the exceptional performance achieved by our pre-fusion strategy. It leverages an attention module to remove irrelevant information and enables the completion module to generate more accurate classifications that are unaffected by sparse or incorrect data. In contrast, the in-fusion and post-fusion strategies tend to postpone the aggregation of information, resulting in inferior performance due to erroneous information. We believe that our pre-fusion strategy has the potential to address issues of uncertainty and incorrectness in future work.

\section{Conclusion and Limitations} 

In this paper, we tackled the challenging problem of traversability assessment for navigation. To achieve this, we proposed a novel framework called FASTC, which achieves end-to-end traversability classification of local regions in LIDAR point clouds by learning point features and encoding them into feature maps in the form of pseudo-images using a Pillar Feature Extraction Network. An attention mechanism is employed to aggregate multi-features and a traversability completion module architecture is used to generate traversability BEV maps. The experiments showed that FASTC outperforms existing methods in terms of both classification accuracy and computational speed on the SemanticKITTI (on-road) and RELLIS-3D (off-road) traversability datasets.

%------------------------------ Table ---------------------------------------
\begin{table}[!t]
\centering
\resizebox{\linewidth}{!}
{
\begin{tabular}{c|cc|cc}
\hline
            & \multicolumn{2}{c|}{SemanticKITTI} & \multicolumn{2}{c}{RELLIS-3D} \\
            & mIoU(\%)         & mAcc(\%)        & mIoU(\%)      & mAcc(\%)      \\ \hline
pre-fusion  & 54.4             & 69.38           & 68.6          & 83.2          \\
in-fusion   & 50.1             & 65.12           & 65.0          & 80.2          \\
post-fusion & 46.3             & 62.1            & 65.1          & 79.6          \\ \hline
\end{tabular}
}
\caption{ \textbf{Effectiveness of different fusion strategies:} Our results demonstrate that pre-fusion strategy produces superior results on both SemanticKITTI and RELLIS-3D datasets.}
\label{tab3}
\end{table}
%---------------------------------------------------------------------

\textbf{Limitations}  We assume neglecting translation and rotation along the z-axis to simplify the differentiable warping process. However, this assumption may not hold for extremely uneven terrain. Additionally, our network lacks memory modules, leading to limited long-term memory capability in the temporal dimension and incomplete recollection of trajectories in certain areas. We plan to address these shortcomings in future improvements.

\bibliography{ecai}
\end{document}